# ForamViT-GAN: Exploring New Paradigms in Deep Learning for Micropaleontological Image Analysis

*Ivan Ferreira-Chacua[1*] and Ardiansyah Koeshidayatullah[1,2*]*

[1] Department of Geosciences, College of Petroleum Engineering and Geosciences, King Fahd University of Petroleum and Minerals, Saudi Arabia

[2]Center for Integrated Petroleum Research, Dhahran, Saudi Arabia*Corresponding author: g202211300@kfupm.edu.sa; a.koeshidayatullah@kfupm.edu.sa

**Abstract**
Micropaleontology in geosciences focuses on studying the evolution of microfossils (e.g., foraminifera) through geological records to reconstruct past environmental and climatic conditions. This field heavily relies on visual recognition of microfossil features, making it suitable for computer vision technology, specifically deep convolutional neural networks (CNNs), to automate and optimize microfossil identification and classification. However, the application of deep learning in micropaleontology is hindered by limited availability of high-quality, high-resolution labeled fossil images and the significant manual labeling effort required by experts.

To address these challenges, we propose a novel deep learning workflow combining hierarchical vision transformers with style-based generative adversarial network algorithms to efficiently acquire and synthetically generate realistic high-resolution labeled datasets of micropaleontology in large volumes. Our study shows that this workflow can generate high-resolution images with a high signal-to-noise ratio (39.1 dB) and realistic synthetic images with a Frechet inception distance similarity score of 14.88. Additionally, our workflow provides a large volume of self-labeled datasets for model benchmarking and various downstream visual tasks, including fossil classification and segmentation.

For the first time, we performed few-shot semantic segmentation of different foraminifera chambers on both generated and synthetic images with high accuracy. This novel meta-learning approach is only possible with the availability of high-resolution, high-volume labeled datasets. Our deep learning-based workflow shows promise in advancing and optimizing micropaleontological research and other visual-dependent geological analyses.

## 1. Introduction

Throughout geological history, foraminifera represent one of the most exceptionally diverse groups of marine microfossils, with an estimated number of current species between 8,966 and an estimated number of 40,888 fossil species in the geological record[1]. This accounts for approximately 2% of all animal species from the Cambrian to the present. (Sen, 2003). The size of fossil foraminifera is very diverse ranging from less than 100 microns to 20 centimeters and their shell can be made up of diverse compositions, such as calcite, aragonite, agglutinated particles, and other organic compounds. In foraminifera, factors such as their cosmopolitan nature and evolutionary diversification make them of particular interest to provide a paleontological and stratigraphic record, which is of significant value to carry out

---

[1] https://www.marinespecies.org/foraminifera/

biostratigraphic correlations and paleoenvironmental interpretations (Jones, 2014; Sen, 2003).

In both ancient and modern environments, the relative abundance of specific species and their corresponding morphometric characteristics are used as a proxy for (paleo) temperature, (paleo) oxygen concentration, and (paleo) oceanic salinity and paleoproductivity. In addition, foraminifera are often used as the building block to define biofacies for paleo-bathymetric studies, as an aid in the characterization of sedimentary sub-environments. With the progressive change of different macro- and microfossils throughout the history of the Earth, especially planktonic foraminifera being utilized as markers on the geological time scale and the occurrence of specific events in the stratigraphic record (Jones, 2014; Marchant et al., 2020; Sen, 2003).

Detailed identification of both species and morphotypes and producing high-quality photomicrographs of foraminifera have been primarily dependent on the availability of high-end equipment such as advanced stereomicroscopes and high-resolution scanning systems. Although some techniques have become staple in research institutes, some high-end equipment for digitizing specimens at high resolutions not widely accessible to the geoscientific community. This is exacerbated by the expertise needed to perform species and genera classification. All of this leads to an issue with standardization across various laboratories and institutions, which severely limits the reproducibility of such classification and its accessibility to non-experts. As a result, there is an urgent need to develop an efficient, automated approach or workflow for improving the resolution of microfossil images and obtaining labeled datasets without the need for high-end equipment. Furthermore, the widespread implementation of robust deep learning models for foraminifera classification and morphological diversity distillation using advanced computer vision technology, particularly deep convolutional neural networks, has yet to be fully investigated.

In the current literature, there are numerous works that leverage computer vision technology to classify and segment microfossil specimens (Beaufort & Dollfus, 2004; Marchant et al., 2020). From seminal implementations in which the algorithms used did not achieve human accuracy (Beaufort & Dollfus, 2004; Culverhouse et al., 1996; S. Liu et al., 1994) to recent works in which the algorithms exceed human accuracy and speed when classifying microfossils (Marchant et al., 2020; Pedraza et al., 2017). Recently, the use of Deep Convolutional Neural Networks (CNNs) has been notable in this research corpus, with CNNs having several advantages for these tasks given the reduced need for feature engineering, the scalability to larger datasets, and exceptional ability to process grid-like data (Carvalho et al., 2020; Ho et al., 2023; Johansen et al., 2021; Koeshidayatullah et al., 2020; Marchant et al., 2020; Mitra et al., 2019). This is further supported by the accessibility to powerful pretrained architectures as backbones for a starting point during the training of new models (Ho et al., 2023; Koeshidayatullah et al., 2020; Koeshidayatullah et al., 2022; Pires de Lima & Duarte, 2021; Shoji et al., 2018).

In geosciences, recent advances of Generative Adversarial Networks (GANs) enable geoscientists to generate additional synthetic data as an additional augmentation technique to conventional ones, assisting in the improvement of machine learning model performance (Ferreira et al., 2022; Wang & Perez, 2017). In addition, GANs are effective at balancing the distribution of data within a particular geological dataset, ensuring a better representation, and reducing the risk of bias during training (Ferreira et al., 2022; Koeshidayatullah, 2022; Abdellatif et al., 2022).

In recent research trends, Vision Transformers (ViT) have emerged from the Natural Language Processing (NLP) corpus as a powerful technique for tackling visual tasks (Vaswani et al 2017; Dosovitskiy et al, 2020). Unlike traditional CNNs, ViTs employ a transformer architecture to address both global and local relationships in an image, resulting in more effective feature extraction and representation. Image Super-Resolution (SR) is a field of Computer Vision that focuses on enhancing low-resolution images, making them more visually appealing and informative. Recent improvements in CNN and ViT-based architectures have become the state-of-the-art for upscaling and restoring images (Liang et al, 2021; Lin et al, 2022). This application could prove useful in preserving fine-grained details and textural information for micropaleontological image analysis.

The primary goals of this research were to create and suggest new approaches to traditional microfossil image scaling methods, and to investigate the application of end-to-end deep learning for enhancing the quality and accurately representing the morphological diversity of foraminifera images. Ferreira et al. (2022) successfully showcased this in their study using petrographic datasets (Fig. 1a and b). Such tools have the potential to expand the variety of micropaleontological datasets and provide synthetic digital counterparts for confidential data.

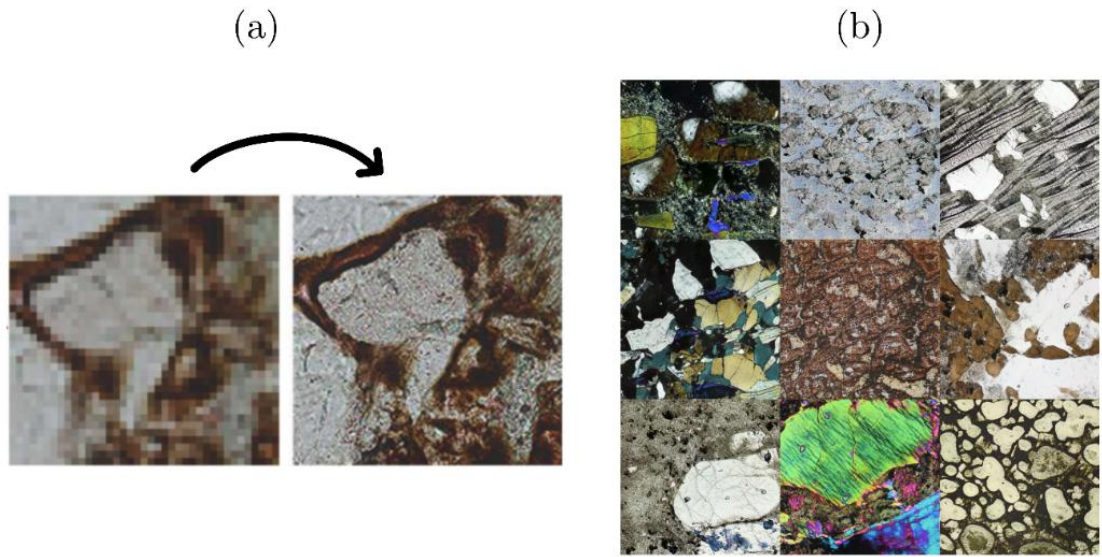

Figure 1: Examples of the use of deep learning architectures in petrography (a) super-resolution imaging of sandstones (Y. Liu et al., 2020) (b) generation of thin sections (Ferreira et al., 2022).

2. **Methodology**

### 2.1. Dataset

In this study, three open-source machine learning-ready databases of foraminifera are used. These datasets were curated and preprocessed before being used in classification tasks using CNNs (Marchant et al., 2020). All three sources are being preprocessed further in preparation for the implementation of the workflow proposed in this study to generate a high-resolution and realistic synthetic micropaleontological dataset.

The first source is the Endless database Forams which is a compilation of more than 34000 foraminifera (Fig. 2a; Hsiang et al., 2019). The second dataset is obtained from the set MD02-2508 compiled by the RV Marion Dufresne oceanography mission MD126 MONA during 2002 in the Northeast Pacific Ocean (Fig. 2b; Marchant et al., 2020). Lastly, the MD97-2138 dataset, these images were collected from the IHPIS mission of the RV Marion Dufresne to analyze the last climatic cycle in sediment cores. (Fig. 2c; Marchant et al., 2020).

These databases are selected for their previous application in training a CNN for classification, as well as for having numerous foraminifera specimens and samples. Furthermore, the datasets vary in terms of backgrounds, illumination, fragmentation of the specimen and number of foraminifera per species. These features are useful for robustly training deep learning model by presenting different potential variations and conditions of foraminifera images acquired from real-world datasets which would help deep learning models in realistically and aesthetically restoring and generating foraminifera dataset.

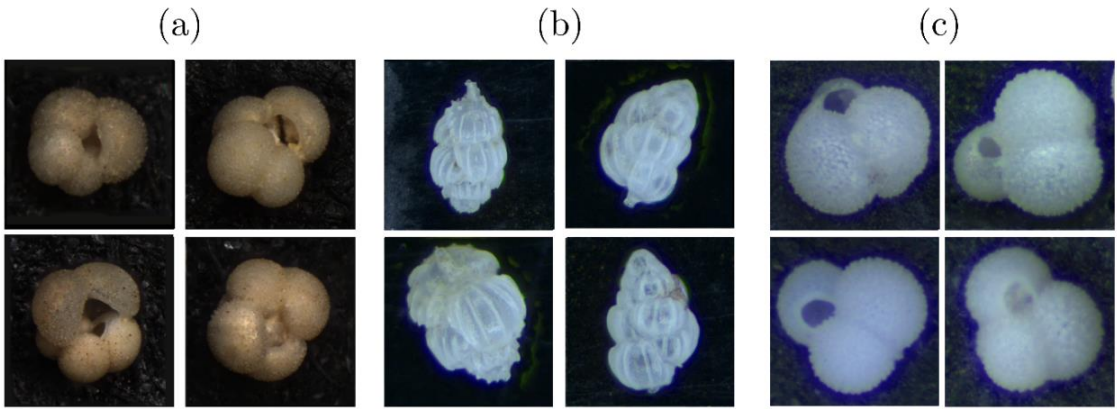

Figure 2: Image samples from the three databases used (a) *Globigerina bulloides from* Endless Forams, (b) *Uvigerina peregrina* from MD02-2508 (c) Neogloboquadrina *dutertrei* from MD97-2138. Selected from the compilation by Marchant et al. 2020.

### 2.2. Transformer-based Image Super Resolution

For the multiscale image restoration task (i.e., image super-resolution), 3,563 specimens are selected from the three databases that meet the condition of having a dimension greater than 800x800 px. This threshold was applied to ensure that the datasets are both above high-definition quality (720px) and has enough samples in the datasets to train the models successfully. The collected images are divided into a training set with 3,263 images and a test set with 300 randomly selected images.

The images were reduced to half, a quarter and an eighth of their original resolution of 800px$^2$, using a Lanczos interpolation kernel, in attempt to preserve the best possible features of the image to be reconstructed. In such a case, the dimensions to be restored are from (i) 400px$^2$ to 800px$^2$ (2x time super-resolution task); (ii) 200px$^2$ to 800px$^2$ (4x time super-resolution task); and 100px$^2$ to 800px$^2$ (8x time super-resolution task). The algorithms were trained until reaching 50000 iterations for comparison, training was done using 25 hours of GPU with an NVIDIA RTX 5000 of 16Gb of video RAM in a Cloud Linux environment, with eight CPU cores and 30 Gb of RAM.

For the foraminifera microphotograph image super-resolution task, our study adopted the SwinIR (Shifted *Windows Transformers for Image Restoration)* (Fig. 3; Liang et al., 2021) architecture which is an extension of the Swin transformer (Z. Liu et al., 2021) algorithm. This architecture is based on the concept of hierarchical neural attention mechanism which has positioned itself as the state-of-the-art deep learning algorithm across standardized datasets used in computer vision tasks.

This trend, combined with their combination with CNNs for architectural improvements, has been seen since the original implementation of the transformers for vision tasks, Vision Transformers (ViT) (Dosovitskiy et al., 2020). With SoTA results, these architecture implements the recent Shifted Window Transformer (Swin) (Z. Liu et al., 2021) applied to image restoration and super-resolution, this architecture uses stacked residual Swin transformer blocks that are windows using a shifted window attention coupled with convolutional layers for further feature extraction (Fig. 3). These allows to achieve better results than current methods by both reducing the number of parameters needed and capturing long-range relationships in the image, allowing in contrast with approaches solely based in CNNs. (Liang et al., 2021; Z. Liu et al., 2021).

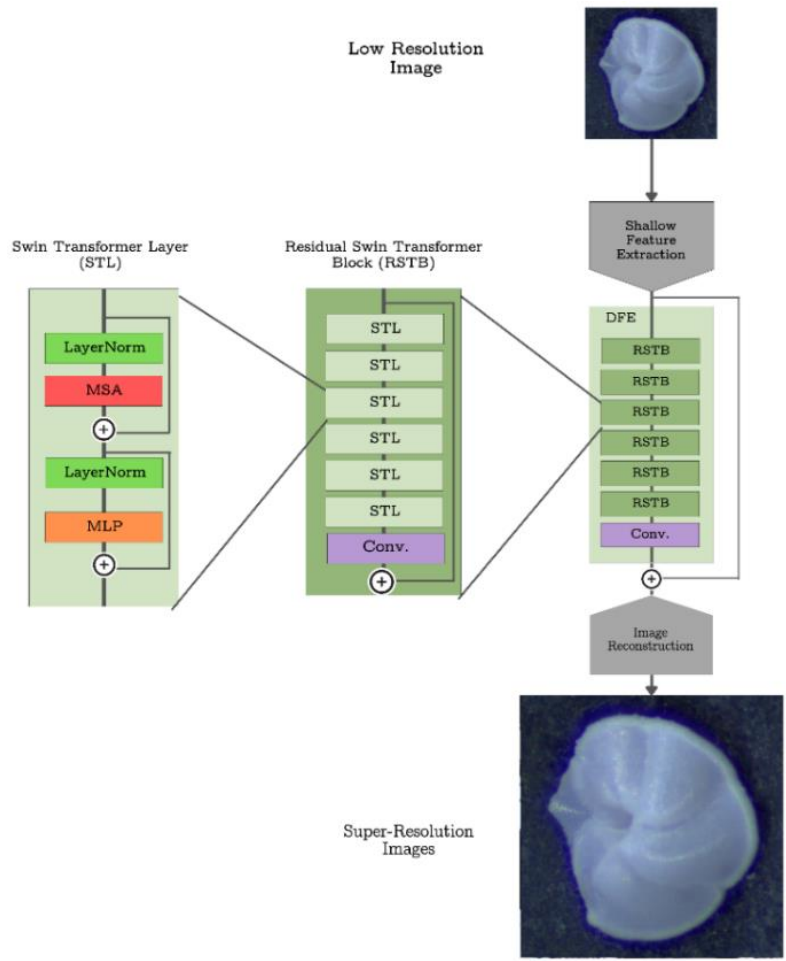

Figure 3: Schematic of the adopted SwinIR architecture used to extract features and upscale images (a) Low-resolution image input to the model (b) Generalized schematic of SwinIR (c) Image upscaled using super-resolution (d) RSTB block, *Residual Swin Transformer Block* (e) Transformer Layer Swin, *Swin transformer Layer*. MSA: Multi-head Self-Attention. MLP: Multi-Layer Perceptron, LayerNorm: Layer Normalization. Modified from (Liang et al., 2021).

### 2.3. Style-based GAN for Image Generation

The generative model was trained by selecting the nine species with the most images (Globigerina *bulloides*, Globigerinita *glutinata*, Globigerinoides *ruber*, Globigerinoides *sacculifer*, Globorotalia *inflata*, Globorotalia *menardii*, Globorotalia *truncatulinoides*, Neogloboquadrina *incompta*, and Neogloboquadrina *pachyderma*). In addition, these species were selected because they share some similarities and distinct characteristics across species simultaneously. A total of 18166 images taken at a resolution of $256px^2$ to obtain the largest number of images and satisfy the requirement of minimum dataset to properly train generative adversarial network ($10^4$-$10^5$ images (Goodfellow et al., 2014; Karras, Aittala, et al., 2020; Karras et al., 2018; Karras, Laine, et al., 2020). Another training was conducted using a 512px$^2$ upsampling of this same dataset was conducted to further experiment with the associated latent space at that scale and evaluate the convergence of the model during training with different dataset resolutions. The idea for this, is to combine the training of GANs and Super-resolution for image generation at higher resolutions with limited data and reduced training time (Fig. 5).

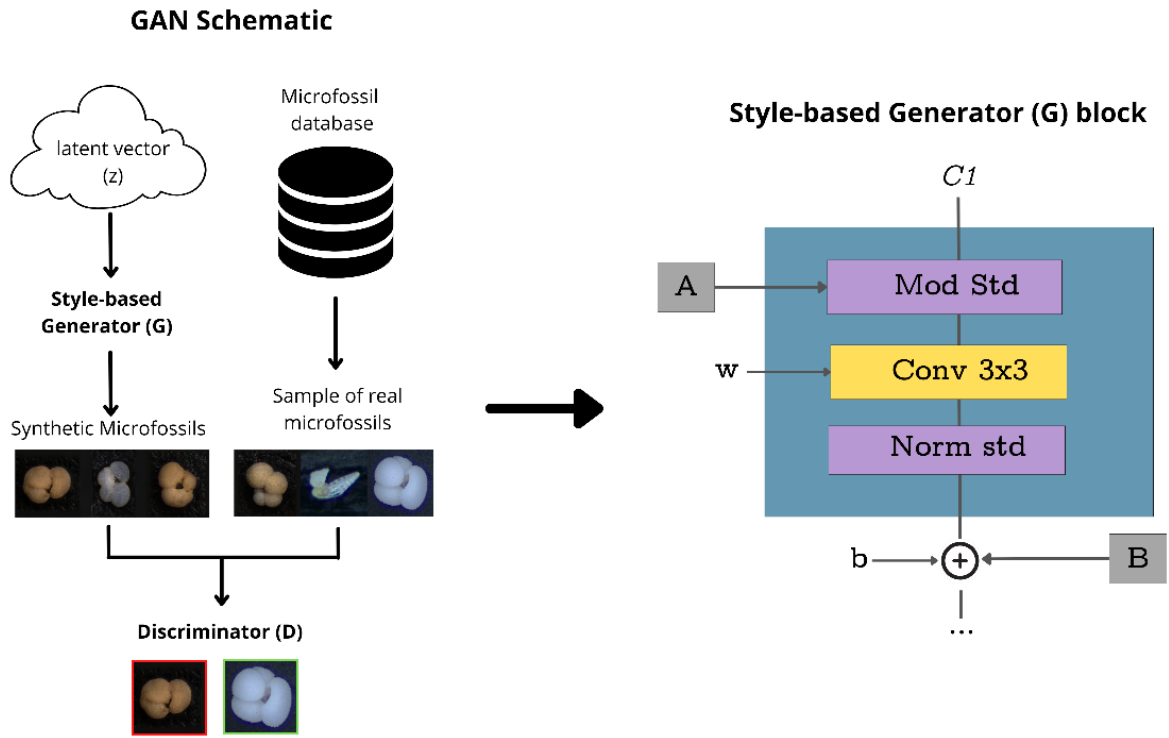

Figure 4: On the left, diagram of a GAN and how it works by taking a latent vector (z), that can be random or conditioned, and a sample from the real dataset. It then learns to replicate this image distribution using a style-based generator (G). These generated images then classified from the real ones using a discriminator (D). The goal of the two neural networks competing in a minimax game in which the goal of G is to deceive D and the goal of D is to discriminate the real samples from the synthetic ones generated by D. On the right, the main Style block used for style extraction on given image datasets by StyleGAN2, with modulation, convolution, normalization, A – Affine transforms, B – Noise broadcast operation, w - learned weights and b – biases (Karras, Aittala, et al., 2020; Karras, Laine, et al., 2020).

Various architectures take advantage of the zero-sum game between two neural networks, which constitutes the core of Generative Adversarial Nets (Goodfellow et al., 2014). Recently, these architectures have been able to deceive the human eye for both face recognition (Lago et al., 2021; Nightingale & Farid, 2022) and even for recognition of specific datasets in geology (Ferreira et al., 2022a; Ferreira & Koeshidayatullah, 2022). In this study, the style-based GAN architecture (StyleGAN2) was adopted and implemented to reconstruct and generate realistic synthetic foraminifera images, (Fig. 4; Karras, Aittala, et al., 2020; Karras et al., 2018; Karras, Laine, et al., 2020). This selection was based on the *Frechet Inception Distance (FID) Score* achieved by this architecture and its variations with the StyleganXL being the current state of the art for image generation tasks (Karras et al., 2021; Sauer et al., 2022). Among other unconditional- and style-based GAN models, the StyleGAN2 architecture has a more robust and flexible implementation that allows experimenting with the model and its associated latent space, allowing not only to generate but also to model synthetic images. For such a case, the algorithm extracts styles of interest in the set of images from which it is trained. These styles can be interpreted in faces as the hair color or the skin tone (Karras et al., 2019), while for petrographic datasets it translates into grain size and color of minerals in the thin section (Ferreira et al., 2022a).

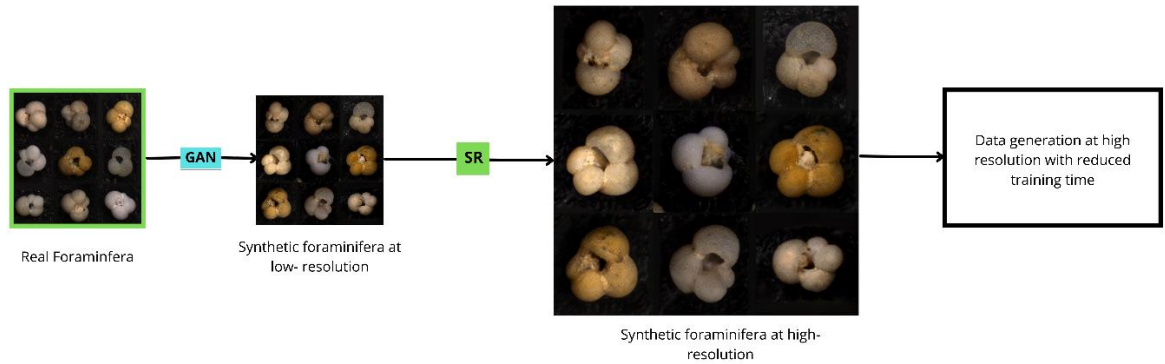

Figure 5: The proposed workflow for coupling GANs with Super-Resolution algorithms to train GANs at lower resolution and upscale the generated images.

### 2.4. Metrics

To evaluate the super-resolution model, we use Peak Signal-To-Noise ratio, PSNR (eq. 1), as a standard metric for the evaluation of super-resolution models and scaling algorithms. This metric has been calculated in the literature for both standard computer vision datasets (Liang et al., 2021; Lin et al., 2022) and in datasets of geological interest (Bizhani et al., 2022; Niu et al., 2020). In addition, this metric calculates the logarithm of the ratio between the maximum value of a signal, 255-pixel value for a grayscale image, and the mean squared error of the image (MSE), eq. 1. In this metric, the higher the value indicating a better the reconstruction of the image, with values greater than 40 is expected from standard image compression algorithms, and undefined when the images are the same, as the MSE goes to zero (refer to eq.1). Values obtained in the literature for geological datasets ranging from 25 to 45 PSNR for different image datasets, from petrographic to micro-computed tomography images (Bizhani et al., 2022; Y. Liu et al., 2020; Niu et al., 2020).

$$PSNR = 10 \times \log_{10}\left(\frac{255^2}{MSE}\right) \quad (1)$$

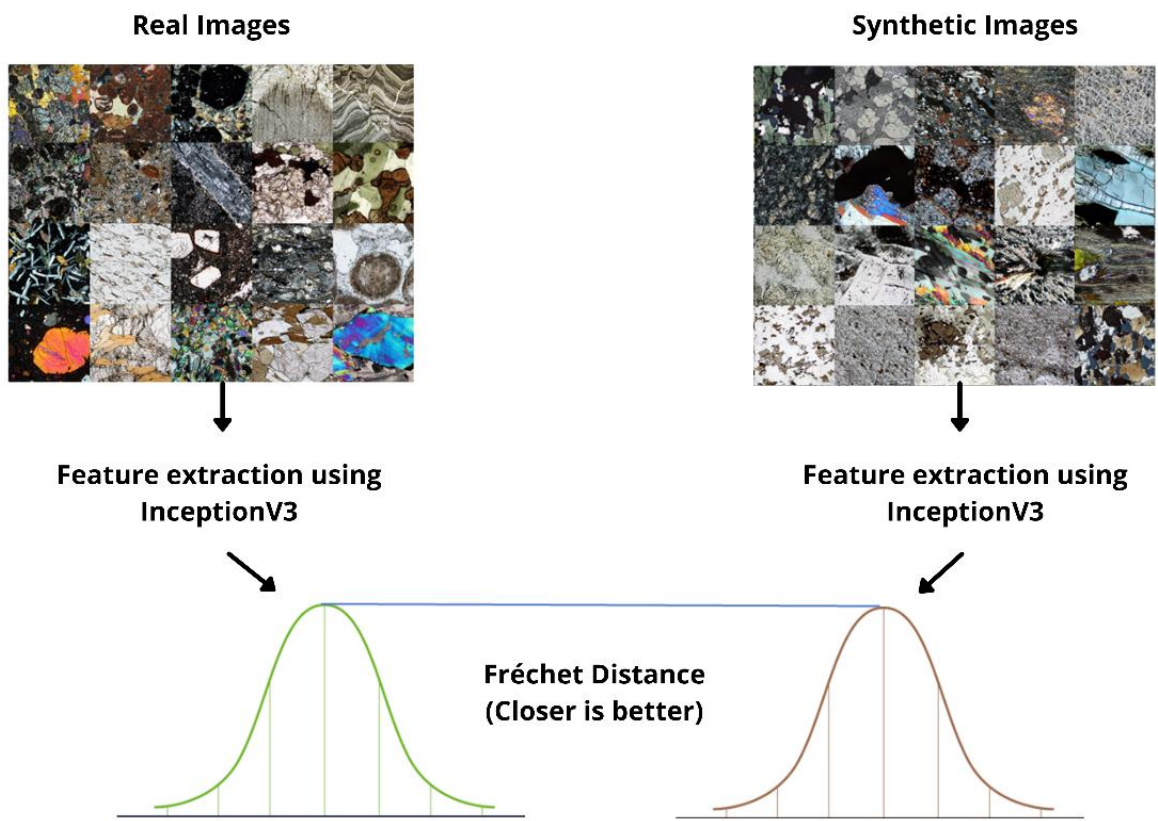

Figure 6: Flowchart to calculate the FID score, taken from (Ferreira et al., 2022).

The generative algorithm is evaluated using the FID Score using InceptionV3 (Heusel et al., 2017; Szegedy et al., 2015). For this metric, the images are sampled, real and fake, and sent as input to the InceptionV3 architecture (Szegedy et al., 2015). The activation results of the last layers of this classification architecture are calculated for both populations, generating a distribution for the synthetic and real images. The Fréchet distance between these two distributions is calculated, the smaller it is, the closer the two distributions are, i.e., the more the synthetic images resemble the real population (Fig. 6). This is a standard metric used for evaluation of generative models, FID score values below 5 expected in face generation using $10^5$ image datasets and values around 10 obtained in generation of synthetic petrographic images (Ferreira et al., 2022b; Heusel et al., 2017; Karras, Aittala, et al., 2020).

## 3. Results

### 3.1. Super-resolution algorithm

The purpose of utilizing the SwinIR architecture was to train it and compare its performance to traditional interpolation methods, as detailed by Gonzalez & Woods (2002) and Liang et al. (2021). The best PSNR value within 50000 iterations of each of the models (2×,4×,8×) is chosen and compared, for the case of the 4× scale, it is also compared with other conventional algorithms, (Fig. 7). The Average PSNR values obtained for the test set in each of the trainings shown in Table 1, together with the iteration of the model in which it was obtained.

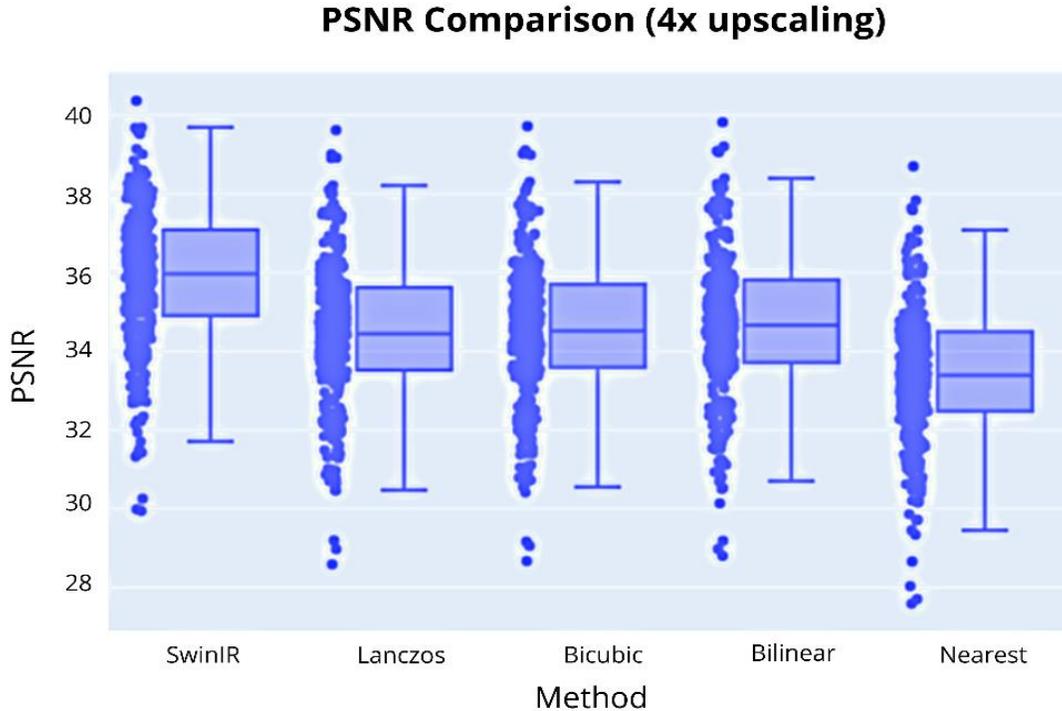

Figure 7: PSNR results for a 4x upscaling (200px$^2$ to 400px$^2$). Comparison between SwinIR and the interpolations using the Lanczos kernel, bicubic bilinear and Nearest Neighbors interpolation.

Table 1: PSNR values obtained for each scaling task and the iteration in which it was reached.

| Scaling | Low resolution (px$^2$) | High Resolution (px$^2$) | PSNR value | Iteration |
|---|---|---|---|---|
| 2× | 400 | 800 | 39.12 | 45000 |
| 4× | 200 | 800 | 35.81 | 50000 |
| 8× | 100 | 800 | 33.29 | 35000 |

In addition, a visual comparison is made of foraminifera and its magnifications both done by SwinIR and by bicubic interpolation and nearest neighbors interpolation (Fig. 8). The results showed that SwinIR outperformed these traditional approaches, demonstrating its potential to surpass traditional methods for image restoration.

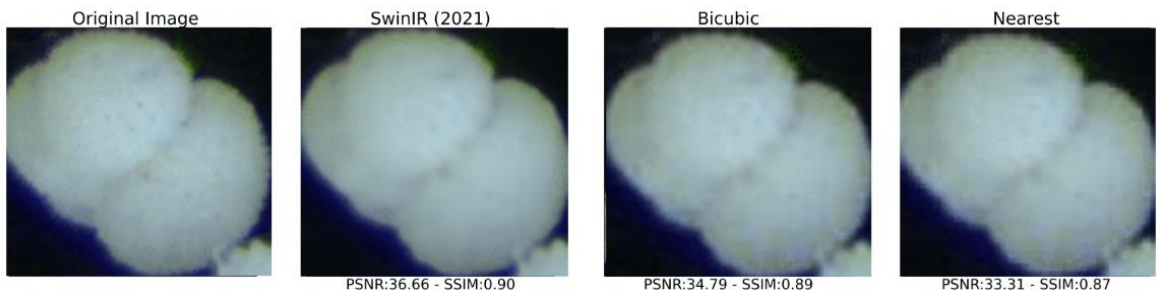
Figure 8: 4x magnification of foraminifera specimens, PSNR: Peak Signal-to-Noise Ratio SSIM: Structural Similarity Index.

### 3.2. Generative algorithm

The StyleGAN2 architecture was also trained to generate synthetic foraminifera specimens and explore their associated latent space (Karras, Aittala, et al., 2020; Karras, Laine, et al., 2020). The performance of the generative model was evaluated both quantitatively, with the FID score (Fig. 9), and qualitatively with a visual inspection of the generated foraminifera (Fig. 10). This was done with the aim of ensuring that the model converged (Fig. 9), as well as that the model did not collapse generating a single foraminifera specimen (Fig. 10). The FID score achieved for the generator at the species level was 14.88 and the score at the unconditioned level is 32.32, observing that the model has a better performance when conditioned to generate specific foraminifera species than it does when it is unconditioned.

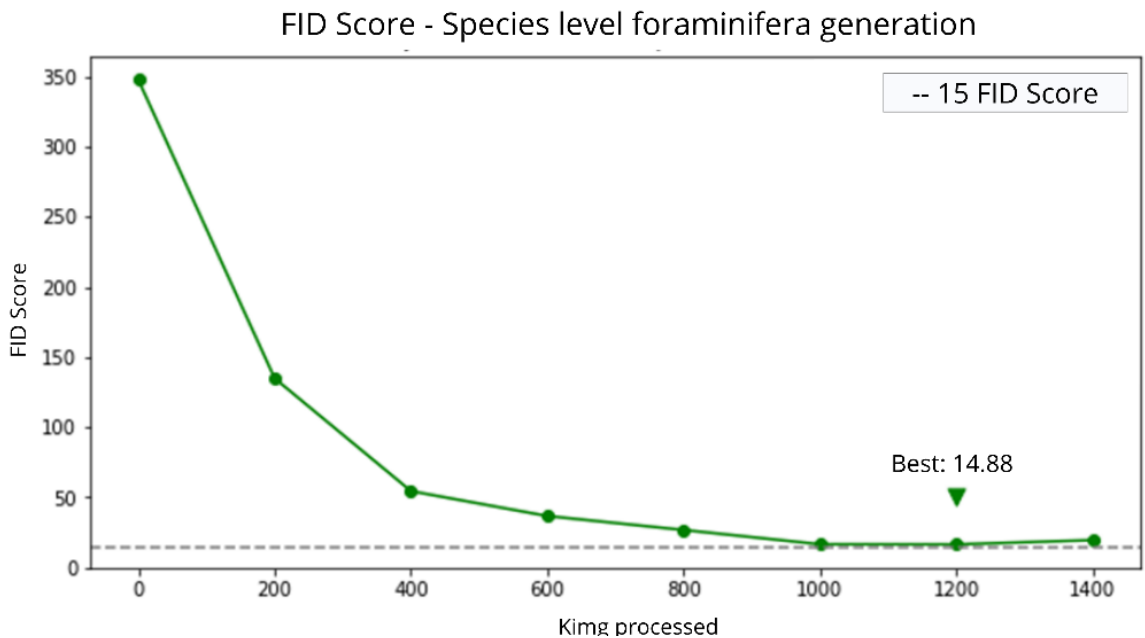

Figure 9: FID score for foraminiferal species, indicating the best score obtained, 14.88.

### 4. Discussion and Conclusions

In addition to the fossil classification and segmentation tasks, the current development of deep learning architectures in using generative adversarial networks (GAN) and Vision Transformers (ViT) to automatically generate and restore geological image datasets has fueled the interests to explore and unravel the potential of these models in geosciences. One of the major tasks is image super resolution (i.e., upscaling image resolution with deep learning) which is instrumental in achieving accurate classification of various geological analysis. Hence, it is highly important to devise a workflow that could bring low-resolution images or images with a considerable amount of noise to a better quality which would directly translate into saving resources when obtaining datasets. In addition, this allows legacy datasets to be re-used and accessible for newer algorithms. On the other hand, generating synthetic data allows for replicating both aesthetic and statistical characteristics of a set of geoscientific datasets, allowing data to be modeled, visualized, and augmented for subsequent geological workflows (Bizhani et al., 2022; Ferreira et al., 2022a; Y. Liu et al., 2020; Mosser et al., 2017; Nanjo Takashi & Tanaka Satoru, 2020; Niu et al., 2020).

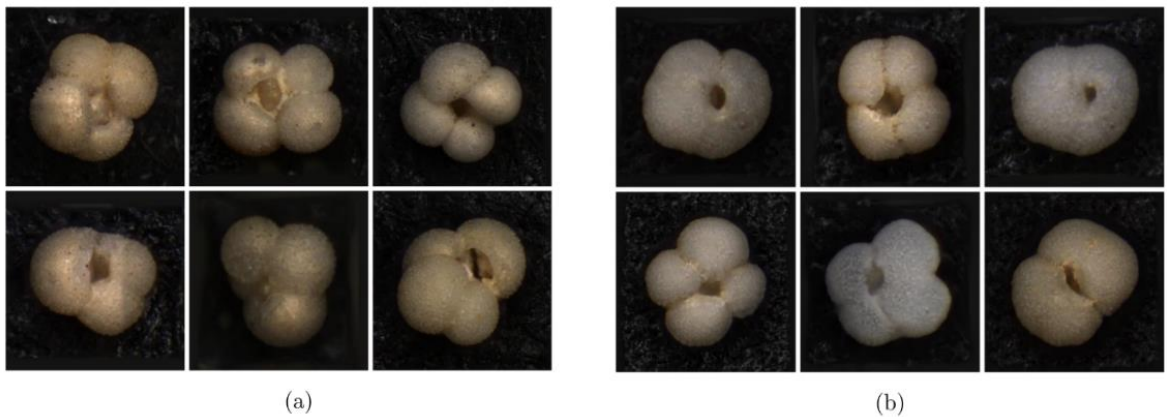

Figure 10: Generation of species-specific foraminifera (a) Real foraminifera of the species *Globigerina bulloides* (b) Generated specimens of *G. bulloides.*

This work seeks to find alternatives to classical microfossil image scaling algorithms, managing to train for the first time a specific architecture, SwinIR, for this purpose, its performance being better than classical interpolations such as Lanczos, bicubic and bilinear, when compared using a standard metric such as PSNR. This demonstrates a potential of this type of architecture to surpass traditional approaches for image restoration, even when this algorithm is trained with a dataset of both moderate size, 3263 images, by computer vision standards (Deng et al., 2009), while also being of specific scientific importance.

In addition, the StyleGAN2 architecture is trained, opening the possibility of generating foraminifera specimens and experimenting with the associated latent space, obtaining a model that reaches FID scores less than 20, being close to the values reported for other geological datasets, which are capable of being indistinguishable from real images to the trained eye (Ferreira et al., 2022a). These foraminifera were visually compared between some selected species, however, a more detailed analysis with experts in the field is required to know how well it replicates morphotypes between different species. The trained model can generate images of foraminifera at the species level and also, another model is trained for not unconditional generation of foraminifera. A possible use of this type of tools is to increase the diversity of micropaleontological datasets, and the release of confidential information as a synthetic copy of the original dataset.

The model's ability to generate synthetic samples can also be applied to few-shot segmentation tasks, as the process of learning how to generate foraminifera generates weights that can be used for other downstream tasks, such as semantic segmentation (Tritrong et al., 2021). In such a case, the model learns how to generate foraminifera via generating weights that could be translated for other downstream tasks, such as semantic segmentation, an example of this given in Fig 11. For this task, the segmentation of foraminifera chambers was performed by only labeling five samples which allow a Few Shots CNN to learn and segment both the aperture and the contour of a specimen, in this case a of a synthetic *Globigerina bulloides*.

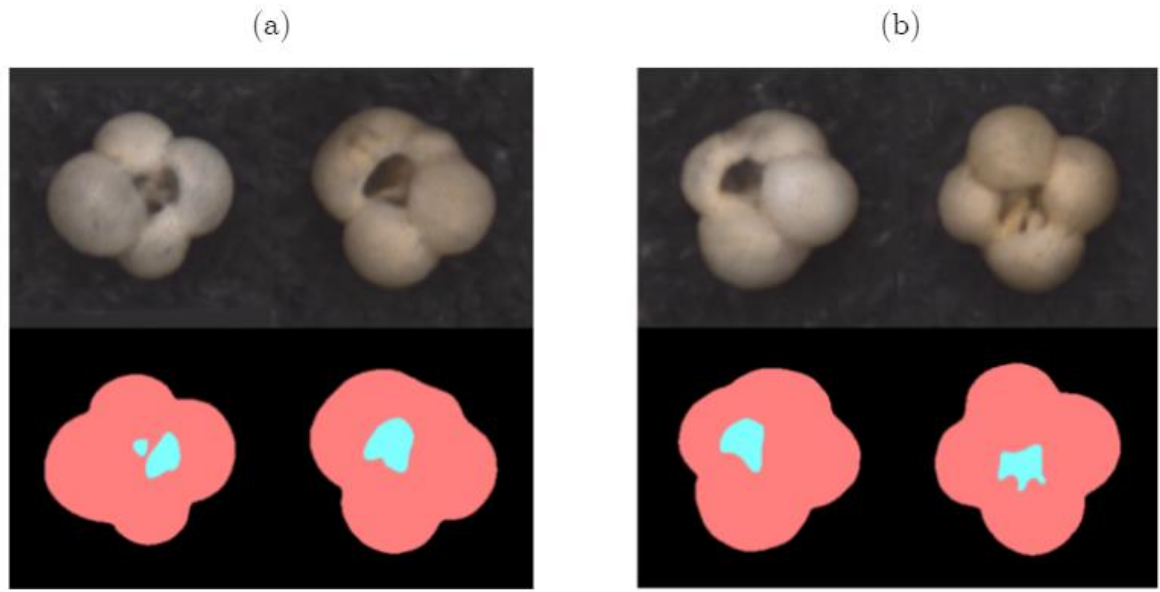

Figure 11: Few-Shot segmentation of *Globigerina bulloides* using the pretrained GAN (a) five specimens labelled (b) model inferences.

**Data availability**
- The data used are available from the cited publication (Marchant et al. 2020), or the preprocessed dataset for upon reasonable request from the authors.
- The models, sample synthetic images,     and images used to evaluate the super-resolution model are located in the drive folder:
  https://drive.google.com/drive/folders/1FGnWC9laJ2F33zDlA3AbvrfNekSNyleR?usp=sharing

**Acknowledgment**

We thank King Fahd University of Petroleum and Minerals for their support through a startup fund grant to A.K (SF21011), we also thank the reviewers of the early versions of this paper for their valuable contributions to the manuscript.

**Bibliography**

Abdellatif, A., Elsheikh, A. H., Graham, G., Busby, D., & Berthet, P. (2022). Generating unrepresented proportions of geological facies using Generative Adversarial Networks. *Computers and Geosciences*, *162*. https://doi.org/10.1016/j.cageo.2022.105085

Beaufort, L., & Dollfus, D. (2004). Automatic recognition of coccoliths by dynamical neural networks. *Marine Micropaleontology*, *51*(1–2), 57–73. https://doi.org/10.1016/j.marmicro.2003.09.003

Bizhani, M., Ardakani, O. H., & Little, E. (2022). Reconstructing high fidelity digital rock images using deep convolutional neural networks. *Scientific Reports*, *12*(1), 4264. https://doi.org/10.1038/s41598-022-08170-8

Carvalho, L. E., Fauth, G., Baecker Fauth, S., Krahl, G., Moreira, A. C., Fernandes, C. P., & von Wangenheim, A. (2020). Automated Microfossil Identification and Segmentation using a Deep Learning Approach. *Marine Micropaleontology*, *158*. https://doi.org/10.1016/j.marmicro.2020.101890

Culverhouse, P. F., Simpson, R. G., Ellis, R., Lindley, J. A., Williams, R., Parisini, T., Reguera, B., Bravo, I., Zoppoli, R., Earnshaw, G., McCall, H., & Smith, G. (1996). Automatic classification of field-collected dinoflagellates by artificial neural network. *Marine Ecology Progress Series*, *139*(1–3), 281–287. https://doi.org/10.3354/meps139281

Deng, J., Dong, W., Socher, R., Li, L.-J., Li, K., & Fei-Fei, L. (2009). ImageNet: A Large-Scale Hierarchical Image Database. *Computer Vision and Pattern Recognition*.

Dosovitskiy, A., Beyer, L., Kolesnikov, A., Weissenborn, D., Zhai, X., Unterthiner, T., Dehghani, M., Minderer, M., Heigold, G., Gelly, S., Uszkoreit, J., & Houlsby, N. (2020). An Image is Worth 16x16 Words: Transformers for Image Recognition at Scale. *ArXiv*, *preprint*. https://github.com/

Ferreira, I., & Koeshidayatullah, A. (2022). Generating and modelling metamorphic thin sections using Generative Adversarial Networks. In R. White (Ed.), *Metamorphic Studies Group - Research In Progress*.

Ferreira, I., Ochoa, L., & Koeshidayatullah, A. (2022a). PetroGAN: A novel GAN-based approach to generate realistic, label-free petrographic datasets. *ArXiv*, *preprint*. http://arxiv.org/abs/2204.05114

Ferreira, I., Ochoa, L., & Koeshidayatullah, A. (2022b). On the generation of realistic synthetic petrographic datasets using a style-based GAN. *Scientific Reports*, *12*(1). https://doi.org/10.1038/s41598-022-16034-4

Gonzalez, R. C., & Woods, R. E. (2002). *Digital image processing* (2nd ed.). Prentice Hall.

Goodfellow, I. J., Pouget-Abadie, J., Mirza, M., Xu, B., Warde-Farley, D., Ozair, S., Courville, A., & Bengio, Y. (2014). Generative Adversarial Networks. *Proceedings of the 27th International Conference on Neural Information Processing Systems*, *2*, 2672–2680. http://arxiv.org/abs/1406.2661

Heusel, M., Ramsauer, H., Unterthiner, T., Nessler, B., & Hochreiter, S. (2017). GANs Trained by a Two Time-Scale Update Rule Converge to a Local Nash Equilibrium. *ArXiv*, *preprint arXiv*(1706.08500v6). http://arxiv.org/abs/1706.08500

Ho, M., Idgunji, S., Payne, J.L. and Koeshidayatullah, A., 2023. Hierarchical multi-label taxonomic classification of carbonate skeletal grains with deep learning. Sedimentary Geology, 443, p.106298. https://doi.org/10.1016/j.sedgeo.2022.106298

Hsiang, A. Y., Brombacher, A., Rillo, M. C., Mleneck-Vautravers, M. J., Conn, S., Lordsmith, S., Jentzen, A., Henehan, M. J., Metcalfe, B., Fenton, I. S., Wade, B. S., Fox, L., Meilland, J., Davis, C. v., Baranowski, U., Groeneveld, J., Edgar, K. M., Movellan, A., Aze, T., … Hull, P. M. (2019). Endless Forams: >34,000 Modern Planktonic Foraminiferal Images for Taxonomic Training and Automated Species Recognition Using Convolutional Neural Networks. *Paleoceanography and Paleoclimatology*, *34*(7), 1157–1177. https://doi.org/10.1029/2019PA003612

Johansen, T. H., Sørensen, S. A., Møllersen, K., & Godtliebsen, F. (2021). Instance segmentation of microscopic foraminifera. *Applied Sciences (Switzerland)*, *11*(14). https://doi.org/10.3390/app11146543

Jones, R. (2014). *Foraminifera and their applications* (1st ed.). Cambridge University Press.

Karras, T., Aittala, M., Hellsten, J., Laine, S., Lehtinen, J., & Aila, T. (2020). Training Generative Adversarial Networks with Limited Data. *ArXiv*, *preprint*(2006.06676v2). http://arxiv.org/abs/2006.06676

Karras, T., Aittala, M., Laine, S., Härkönen, E., Hellsten, J., Lehtinen, J., & Aila, T. (2021). Alias-Free Generative Adversarial Networks. *Arxiv*, *preprint*(2106.12423v4). http://arxiv.org/abs/2106.12423

Karras, T., Laine, S., & Aila, T. (2018). A Style-Based Generator Architecture for Generative Adversarial Networks. *ArXiv*, *preprint arXiv*(1812.04948v3). http://arxiv.org/abs/1812.04948

Karras, T., Laine, S., Aittala, M., Hellsten, J., Lehtinen, J., & Aila, T. (2019). Analyzing and Improving the Image Quality of StyleGAN. *ArXiv*, *preprint arXiv*(1812.04948). http://arxiv.org/abs/1912.04958

Karras, T., Laine, S., Aittala, M., Hellsten, J., Lehtinen, J., & Aila, T. (2020). Analyzing and Improving the Image Quality of StyleGAN. *ArXiv*, *preprint*(1912.04958v2). http://arxiv.org/abs/1912.04958

Koeshidayatullah, A., Morsilli, M., Lehrmann, D. J., Al-Ramadan, K., & Payne, J. L. (2020). Fully automated carbonate petrography using deep convolutional neural networks. *Marine and Petroleum Geology*, *122*. https://doi.org/10.1016/j.marpetgeo.2020.104687

Koeshidayatullah, A., Trower, E.J., Li, X., Mukerji, T., Lehrmann, D.J., Morsilli, M., Al-Ramadan, K. and Payne, J.L., 2022. Quantitative evaluation of the roles of ocean chemistry and climate on ooid size across the Phanerozoic: Global versus local controls. Sedimentology, 69(6), pp.2486-2506. https://doi.org/10.1111/sed.12998

Koeshidayatullah, A., 2022. Optimizing image-based deep learning for energy geoscience via an effortless end-to-end approach. Journal of Petroleum Science and Engineering, 215, p.110681. https://doi.org/10.1016/j.petrol.2022.110681

Lago, F., Pasquini, C., Böhme, R., Dumont, H., Goffaux, V., & Boato, G. (2021). More Real than Real: A Study on Human Visual Perception of Synthetic Faces. *Arxiv*, *preprint*(2106.07226v2). https://doi.org/10.1109/MSP.2021.3120982

Liang, J., Cao, J., Sun, G., Zhang, K., van Gool, L., & Timofte, R. (2021). SwinIR: Image Restoration Using Swin Transformer. *ArXiv*, *preprint*. http://arxiv.org/abs/2108.10257

Lin, Z., Garg, P., Banerjee, A., Magid, S. A., Sun, D., Zhang, Y., van Gool, L., Wei, D., & Pfister, H. (2022). Revisiting RCAN: Improved Training for Image Super-Resolution. *ArXiv*, *preprint*. http://arxiv.org/abs/2201.11279

Liu, S., Thonnat, M., & Berthod, M. (1994). Automatic Classification of Planktonic Foraminifera by a Knowledge-Based System. *IEEE Proceedings of the Tenth Conference on Artificial Intelligence for Applications* .

Liu, Y., Guo, C., Cao, J., Cheng, Z., Ding, X., Lv, L., Li, F., & Gong, M. (2020). A new resolution enhancement method for sandstone thin-section images using perceptual GAN. *Journal of Petroleum Science and Engineering*, *195*. https://doi.org/10.1016/j.petrol.2020.107921

Liu, Z., Lin, Y., Cao, Y., Hu, H., Wei, Y., Zhang, Z., Lin, S., & Guo, B. (2021). Swin Transformer: Hierarchical Vision Transformer using Shifted Windows. *ArXiv*, *preprint*. http://arxiv.org/abs/2103.14030

Marchant, R., Tetard, M., Pratiwi, A., Adebayo, M., & de Garidel-Thoron, T. (2020). Automated analysis of foraminifera fossil records by image classification using a convolutional neural network. *Journal of Micropalaeontology*, *39*(2), 183–202. https://doi.org/10.5194/jm-39-183-2020

Mitra, R., Marchitto, T. M., Ge, Q., Zhong, B., Kanakiya, B., Cook, M. S., Fehrenbacher, J. S., Ortiz, J. D., Tripati, A., & Lobaton, E. (2019). Automated species-level identification of planktic foraminifera using convolutional neural networks, with comparison to human performance. *Marine Micropaleontology*, *147*, 16–24. https://doi.org/10.1016/j.marmicro.2019.01.005

Mosser, L., Dubrule, O., & Blunt, M. J. (2017). Reconstruction of three-dimensional porous media using generative adversarial neural networks. *Physical Review E*, *96*(4). https://doi.org/10.1103/PhysRevE.96.043309

Nanjo Takashi, & Tanaka Satoru. (2020). Carbonate Lithology Identification with Generative Adversarial Networks. *International Petroleum Technology Conference*.

Nightingale, S. J., & Farid, H. (2022). AI-synthesized faces are indistinguishable from real faces and more trustworthy. *Proceedings of the National Academy of Sciences of the United States of America*, *119*(8). https://doi.org/10.1073/pnas.2120481119

Niu, Y., Wang, Y. da, Mostaghimi, P., Swietojanski, P., & Armstrong, R. T. (2020). An Innovative Application of Generative Adversarial Networks for Physically Accurate Rock Images With an Unprecedented Field of View. *Geophysical Research Letters*, *47*(23). https://doi.org/10.1029/2020GL089029

Pedraza, A., Bueno, G., Deniz, O., Cristóbal, G., Blanco, S., & Borrego-Ramos, M. (2017). Automated diatom classification (Part B): A deep learning approach. *Applied Sciences (Switzerland)*, *7*(5). https://doi.org/10.3390/app7050460

Pires de Lima, R. P., & Duarte, D. (2021). Pretraining convolutional neural networks for mudstone petrographic thin-section image classification. *Geosciences (Switzerland)*, *11*(8). https://doi.org/10.3390/GEOSCIENCES11080336

Sauer, A., Schwarz, K., & Geiger, A. (2022). StyleGAN-XL: Scaling StyleGAN to Large Diverse Datasets. *ArXiv*, *preprint*. http://arxiv.org/abs/2202.00273

Sen, B. (2003). *Modern foraminifera* (1st ed.). Kluwer Academic Publishers.

Shoji, D., Noguchi, R., Otsuki, S., & Hino, H. (2018). Classification of volcanic ash particles using a convolutional neural network and probability. *Scientific Reports*, *8*(1). https://doi.org/10.1038/s41598-018-26200-2

Szegedy, C., Vanhoucke, V., Ioffe, S., Shlens, J., & Wojna, Z. (2015). Rethinking the Inception Architecture for Computer Vision. *ArXiv*, *preprint*(1512.00567v3). http://arxiv.org/abs/1512.00567

Tritrong, N., Rewatbowornwong, P., & Suwajanakorn, S. (2021). Repurposing GANs for One-shot Semantic Part Segmentation. *ArXiv*, *preprint*(2103.04379v5). http://arxiv.org/abs/2103.04379

Vaswani, A., Shazeer, N., Parmar, N., Uszkoreit, J., Jones, L., Gomez, A. N., Kaiser, L., & Polosukhin, I. (2017). Attention Is All You Need (Version 5). arXiv. https://doi.org/10.48550/ARXIV.1706.03762

Wang, J., & Perez, L. (2017). The Effectiveness of Data Augmentation in Image Classification using Deep Learning. *ArXiv*, *preprint arXiv*(1712.04621).